%% file: main.tex
\documentclass[11pt]{article}
\pdfoutput=1

\usepackage[final]{acl}

\usepackage{times}
\usepackage{latexsym}

\usepackage[T1]{fontenc}

\usepackage[utf8]{inputenc}

\usepackage{microtype}

\usepackage{inconsolata}

\usepackage{graphicx}
\usepackage{subcaption}
\usepackage{booktabs}
\usepackage{amssymb}
\usepackage{amsmath}
\usepackage{multirow}
\usepackage{tabularx}
\usepackage{bm}
\usepackage{xcolor}
\usepackage[most]{tcolorbox}
\usepackage{pifont}
\usepackage{enumitem}
\usepackage{listings}
\usepackage{algorithm}
\usepackage{longtable}
\usepackage{array}
\usepackage{ragged2e}

\title{VDAR-Router: Adaptive LLMs Routing via \underline{V}erbalized Query \underline{D}ifficulty
\underline{A}nalysis \underline{R}etrieval}

\author{
    Yu-Chien Tang$^{\dagger}$, Jun-Chen Hung$^{\dagger}$, Wen-Chih Peng, An-Zi Yen \\
    Department of Computer Science, National Yang Ming Chiao Tung University, Taiwan\\
    \texttt{tommytyc.cs10@nycu.edu.tw, andyhung.cs13@nycu.edu.tw,}\\ \texttt{wcpeng@cs.nycu.edu.tw, azyen@nycu.edu.tw} \\
}

\begin{document}
\maketitle

\def\thefootnote{$\dagger$}\footnotetext{Both authors contributed equally to this work.}\def\thefootnote{\arabic{footnote}}

\input{sections/00-abstract}
\input{sections/01-introduction}
\input{sections/02-relatedwork}
\input{sections/03-method}
\input{sections/04-experiment-setup}
\input{sections/05-experiment-result}
\input{sections/06-conclusion}

\input{sections/98-limitation}

\section*{Acknowledgments}
This research was partially supported by the National Science and Technology Council, Taiwan, under grants NSTC 114-2221-E-A49-057-MY3, NSTC 114-2639-E-A49-001-ASP, and NSTC 115-2639-E-A49-001-ASP. Additional support was provided by the Research Center for Information Technology Innovation (CITI), Academia Sinica, under the project ``Advancements in Industrial and Financial Large Language Models: Construction and Applications,'' and Delta Electronics under the project ``The Construction and Application of Large Language Models in Intelligent and/or Green Buildings.''

\bibliography{refrence}

\appendix
\input{sections/99-appendix}

\end{document}

%% file: sections/00-abstract.tex
\begin{abstract}
Large language models are increasingly used in practical systems, making efficient model selection important for reducing deployment cost.
LLM routing has emerged as a practical solution for allocating each input query to an appropriate model under a desired cost-performance trade-off.
Existing routing methods often estimate model suitability from the surface semantics or embedding similarity of the input query.
However, such methods may ignore the underlying difficulty of a query, leading to suboptimal routing decisions.
To address the challenge, we propose VDAR-Router, a difficulty-aware retrieval-based routing framework.
For each input query, VDAR-Router first generates an explicit difficulty analysis.
It then retrieves historical examples with similar difficulty profiles.
Based on the retrieved records, it estimates candidate model suitability and selects the model using a reward function that considers both performance and cost.
Experiments on three datasets show that VDAR-Router consistently achieves better cost-performance trade-offs than existing baselines.
These results demonstrate the effectiveness of difficulty-aware retrieval for training-free LLM routing.
Case studies further show that explicit query analysis helps retrieve more relevant examples and supports more reliable routing decisions.\footnote{Our code is publicly available at \url{https://github.com/NYCU-NLP-Lab/VDAR-Router}}

\end{abstract}

%% file: sections/01-introduction.tex
\section{Introduction}

Large language models (LLMs) have evolved rapidly and achieved strong performance across a broad range of tasks, including summarization, code generation, and mathematical reasoning \citep{xue2024decompose, wang2024mmlupro, tian2025template, singh2026openaigpt5card}.
At the same time, LLMs are increasingly embedded in agentic workflows that plan actions, invoke tools, retrieve information, inspect code, and make multi-step decisions \citep{yao2023react, schick2023toolformer, qin2024tooluse, zheng-etal-2024-opencodeinterpreter}, with emerging systems such as OpenClaw\footnote{\url{https://openclaw.ai/}} reflecting this trend in practical applications.
These workflows often require a sequence of heterogeneous model calls, ranging from simple information extraction to complex reasoning or code editing.
However, always invoking the strongest model can quickly exhaust user budgets or service quotas, while relying on cheaper but weak models may lead to poor responses to difficult queries.
As models differ in their architectures, training data, parameter scales, and capability profiles, they exhibit distinct strengths, costs, latency, and response characteristics across tasks \citep{barandoni2024automating}, posing a unique challenge in dynamically identifying the most appropriate model within budget for a variety of user queries.

\begin{figure}[t!]
    \centering
    \includegraphics[width=1\linewidth]{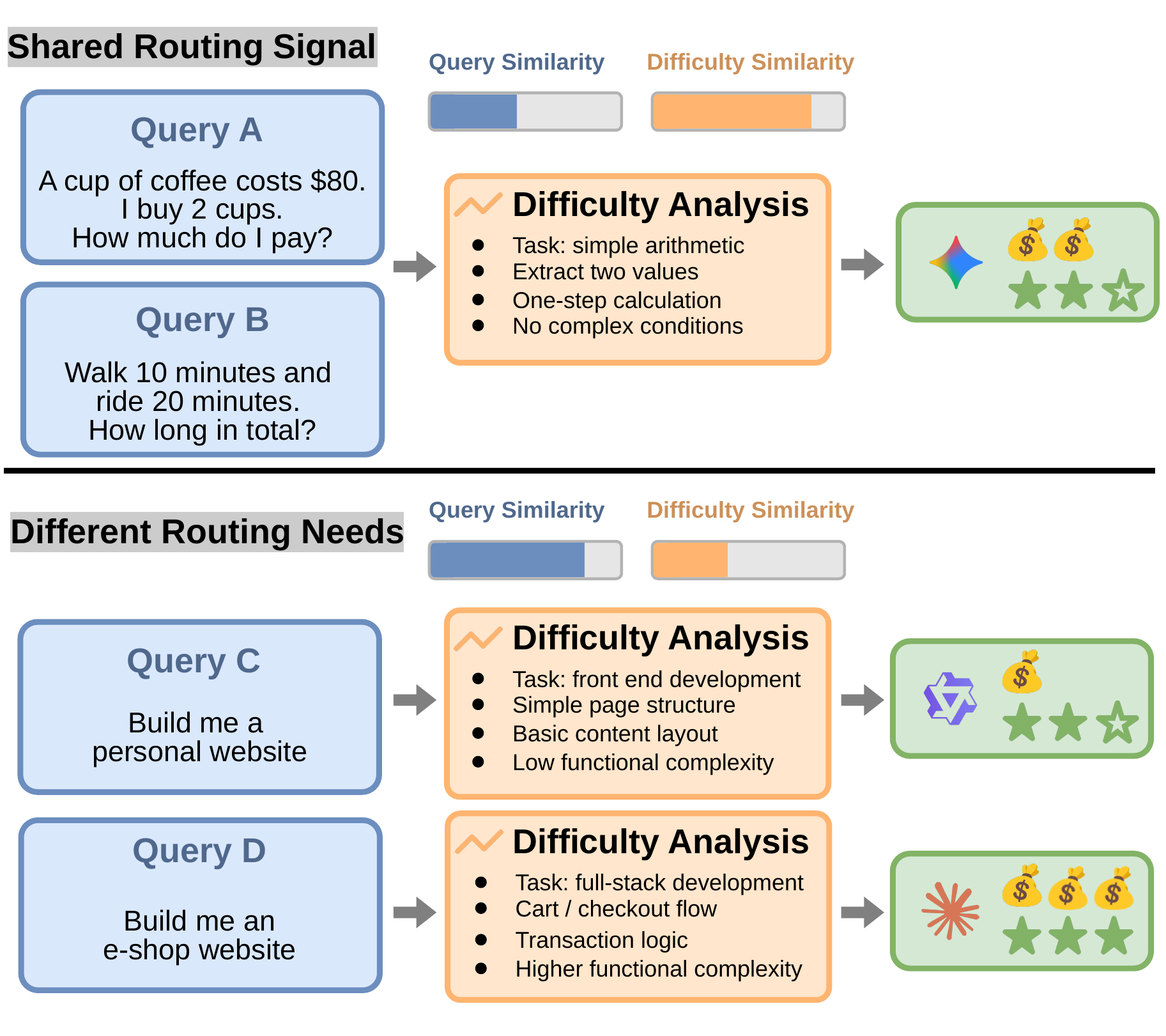}
    \caption{An example of using difficulty analysis to select model instead of using raw query.}
    \label{fig:scenario}
\end{figure}

A growing body of work has studied LLM routing as a way to strike a balance between response quality and inference cost.
One common direction addresses routing in a binary manner, where simpler queries are assigned to a cheap model and more complex queries are routed to a stronger one \citep{chen2024frugalgpt,ding2024hybrid,ong2025routellm}.
Building on this formulation, recent methods train learned routers from labeled examples, preference data, or benchmark performance records to predict the most suitable model for each query \citep{chen2024routerdc,feng2025graphrouter,dai2024costeffectiveonlinemultillmselection,wang-etal-2025-mixllm}.
Other approaches further incorporate retrieval or difficulty estimation, comparing an incoming query with previously observed examples or explicitly modeling query difficulty with similar queries to support routing decisions \citep{stripelis-etal-2024-tensoropera,song-etal-2025-irt}.
Nevertheless, most of these existing methods still heavily depend on the query itself as the primary routing signal.
The router is often expected to learn or retrieve from query representations that mainly encode surface-level semantics, such as the topic, wording, or task category of the input, while semantic similarity may not always reflect the difficulty characteristics relevant to model selection.

Based on these findings, we aim to observe the capability required to answer and the difficulty of the query to produce a fine-grained knowledge of the query.
We hypothesize that if the queries with similar difficulty share similar skills to correctly answer, each model must have similar response quality on these queries, thus providing reliable evidence for model routing.
As illustrated in Figure~\ref{fig:scenario}, two arithmetic queries can look unrelated in raw text, but both require only simple one-step reasoning.
Conversely, two website-generation queries may appear similar in semantics, but one may only require a simple personal webpage while the other demands more complex functionality, planning, and implementation details.
In these cases, if the router understands the performance of each model on queries with similar difficulty analysis, it can better capture the capability ranking of the incoming query and select the most proper one.

In this paper, we propose VDAR-Router, a difficulty-aware retrieval-based framework for LLM routing.
Given an input query, VDAR-Router first analyzes its difficulty characteristics and uses the resulting analysis representation to retrieve historically observed queries with similar routing-relevant challenges.
It then estimates model suitability from the historical performance or preference signals associated with the retrieved examples.
Finally, the routing decision is made by jointly considering response quality and inference cost.
This design allows VDAR-Router to reuse routing evidence across queries with shared difficulty patterns, while distinguishing superficially similar queries that require different model capabilities.

In sum, our contributions are threefold:
\begin{itemize}[leftmargin=*]
\item
We introduce a novel LLM routing approach by incorporating the verbalized diagnosis of query difficulty as the retrieval objectives.
The analysis result can also serve as human-interpretable metadata for each routing decision.

\item
We propose VDAR-Router, a plug-and-play retrieval-based routing framework.
By matching the difficulty characteristics of a test query with historical queries and considering cost to rerank model capability orders, VDAR-Router enables model suitability estimation without requiring any additional training.

\item
Experimental results on three model selection datasets demonstrate that VDAR-Router can effectively recognize the relative model capability rankings, providing a refreshing view for practitioners in LLM routing.
\end{itemize}

%% file: sections/02-relatedwork.tex
\section{Related Work}

\begin{figure*}[t]
    \centering
    \includegraphics[width=1\linewidth]{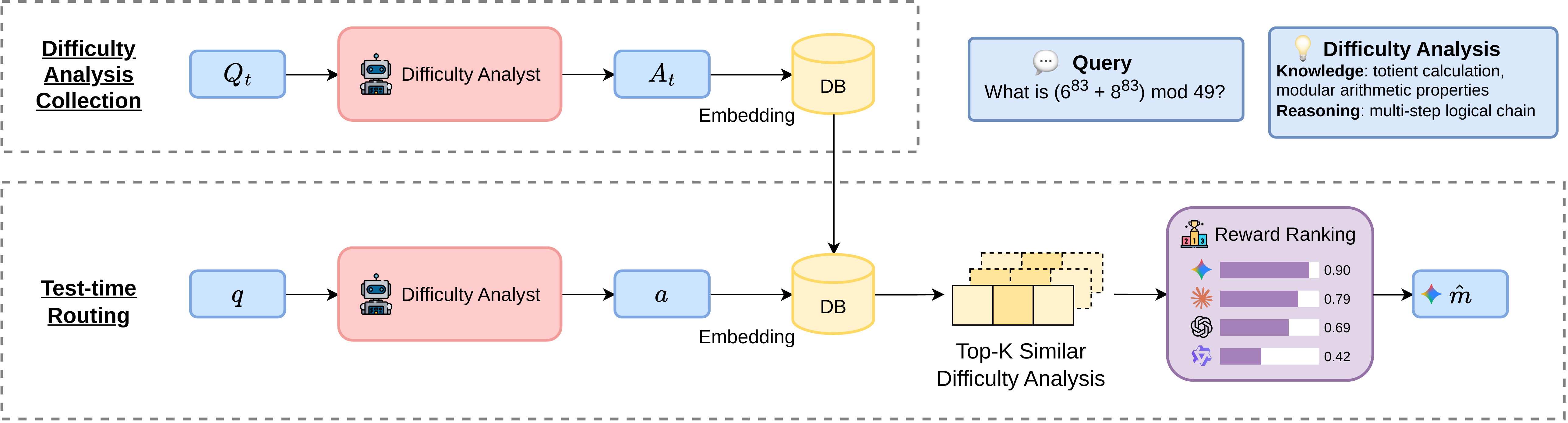}
    \caption{Overview of the VDAR-Router workflow.}
    \label{fig:framework}
\end{figure*}

Efficient LLM routing, which aims to direct an incoming query to the most appropriate model, has recently been explored extensively. 
Early work was mainly motivated by the cost-quality balance in LLM deployment by selecting the answer model from a cheap but weak model and an expensive but strong one \citep{chen2024frugalgpt,ding2024hybrid,ong2025routellm}.
These works established the core intuition of LLM routing: different queries require different levels of model capability, and routing can exploit this heterogeneity to reduce cost.

Recent works build router framework in a supervised manner.
RouterDC \citep{chen2024routerdc} uses dual contrastive learning to learn query and model representations for LLM selection. 
GraphRouter \citep{feng2025graphrouter} represents tasks, queries, and LLMs as a heterogeneous graph and formulates routing as an edge prediction problem. 
Some researchers also leverage bandit-based learning to optimize the routing strategy \citep{dai2024costeffectiveonlinemultillmselection,wang-etal-2025-mixllm}.
However, a common limitation of these methods is their reliance on training an additional neural model and the limiting interpretability of each routing decision, which can complicate real-world deployment.

Another line of research explores the adoption of the retrieval method to assess whether a LLM candidate is capable of answering a query.
\citet{stripelis-etal-2024-tensoropera} employs KNN retrieval and selects the LLM achieving the highest performance on retrieved queries.
IRT-Router \citep{song-etal-2025-irt} shares similar idea, and further applies Item Response Theory (IRT) to estimate the numeric query difficulty and calibrate LLMs performance ranking.
Instead of using query embedding, we conduct retrieval over the embedding of verbalized query difficulty analysis to better recognize the performance of each LLMs on queries with similar capability requirement.

%% file: sections/03-method.tex
\section{Method}

\subsection{Preliminary}
\subsubsection{Problem Formulation}
We formally define a training set of queries $Q_t = \{q_t^1, q_t^2, ...\}$, a set of candidate LLMs, $M = \{m^1, m^2, ..., \}$, and the response performance and cost on $Q_t$ of each LLM.
The goal of router is to learn from $Q_t$ to intelligently select the optimal model $m \in M$ for incoming query $q$, balancing both response quality and cost.

\subsubsection{VDAR-Router Overview}

We present a training-free and easy-to-implement framework, VDAR-Router, to address the LLM routing problem through difficulty-aware query-model matching.
As illustrated in Figure \ref{fig:framework}, an LLM agent \textit{Difficulty Analyst} is designed to determine the proficiency levels across various capability dimensions required for an LLM to correctly answer a given query $q$.
We first offline collect the difficulty analyses of training queries by prompting Difficulty Analyst, and embed the results in a database (Section \ref{sec:diff_agent}).
During routing, after producing the difficulty analysis for incoming queries, we retrieve the most similar difficulty analyses and ranking the candidate models by their performance and cost on the corresponding retrieved queries (Section \ref{sec:runtime}).
In this way, the capability order of each model is accordingly aligned with its answer correctness of queries with similar difficulty level, and thus enables a more adaptive routing decision.

\subsection{Difficulty Analysis Collection} \label{sec:diff_agent}

The goal of the Difficulty Analyst is to estimate the difficulty characteristics of an input query by identifying the capabilities required to solve it.
We hypothesize that explicitly reasoning over the required capability dimensions and their corresponding proficiency levels enables the router to build a more fine-grained understanding of the query, thus improving LLM selection \citep{chen2025symbolic,yu-etal-2025-mexa}.
To this end, we design the Difficulty Analyst, denoted as $\mathcal{M}$, to infer the extent of capabilities necessary for answering a given query.
Inspired by \citet{minaee2024large} and \citet{shi2025inferencedynamics}, we consider seven capability dimensions: reasoning, comprehension, instruction following, agentic capabilities, knowledge retrieval, coding, and multilingual ability.
These dimensions are integrated into the system prompt of $\mathcal{M}$ as analytic targets.
Given an input query $q$, the agent produces a query difficulty analysis:
\begin{equation}
    a = \mathcal{M}(q),
\end{equation}
where $a$ describes the difficulty characteristics and required capabilities of query $q$ across the predefined dimensions.

To collect the difficulty analysis for routing, we apply the same analysis process to each query $q_t^i$ in training set $Q_t$.
The generated difficulty analysis $a_t^i$ is encoded by an embedding model $\mathcal{E}$ and indexed in database for later retrieval.
This design allows VDAR-Router to retrieve examples according to difficulty similarity rather than query similarity.
Consequently, lexically different queries may still be retrieved together for model capability reference when they require similar capabilities, and superficially similar queries with different difficulty profiles can be separated.\footnote{The prompt of Difficulty Analyst is in Appendix \ref{app:difficulty-analysis-prompt}.}

\subsection{Test-time Difficulty-aware Routing} \label{sec:runtime}

At runtime, given an incoming query $q$, the Difficulty Analyst first produces its difficulty analysis $a$.
The generated analysis is then embedded by the same embedding model $\mathcal{E}$ and is used to retrieve top-$k$ similar analyses $a_r$ with similar difficulty characteristics from the database.
The retrieved $a_r$ are then mapped back to the corresponding queries $q_r \in Q_t$ and the response performance of each LLM, where the performance ranking represents the capability order of each model on similar difficulty queries.
However, only selecting the model performed the best on $q_r$ may ignore the impact of inference cost and always seek the strongest but expensive models \citep{jitkrittum2026universal,varangot-reille-etal-2026-generalising}.
In light of this, we propose to rerank the order of models by integrating the cost and calculating reward $R$: 
\begin{equation}
    \label{eq:reward}
    R(m \mid q) = \frac{1}{k} \sum_{i=1}^{k}  \alpha \cdot p_m(q_r^i) - \beta \cdot c_m(q_r^i),
\end{equation}
where $p_m(q_r^i)$ denotes the performance of model $m$ on $q_r^i$, and $c_m(q_r^i)$ denotes the cost of $m$ on $q_r^i$ linearly normalized to $[0,1]$.
The coefficients $\alpha$ and $\beta$ control the trade-off between performance and cost, with $\alpha + \beta = 1$.
Finally, the router selects the candidate model with the highest reward:
\begin{equation}
\hat{m} = \arg\max_{m \in M} R(m \mid q).
\end{equation}

%% file: sections/04-experiment-setup.tex
\section{Experimental Setup}

\input{tables/01-bench-result}

\subsection{Implementation Details}
Across all experiments and data preprocessing, we set random seed to 42, temperature to 0, retrieval size $k$ to 30, and LLMs decoding strategy to greedy for reproducibility. 
All the experiments are conducted on two NVIDIA RTX A6000 GPUs.

\subsection{Datasets}

We evaluate our routing method on three datasets: RouterBench \citep{hu2024routerbench}, LLMRouterBench \citep{li2026llmrouterbench}, and ArenaExpert5K \citep{arena2025arenaexpert5k}.
RouterBench and LLMRouterBench represent regular routing settings, where the performance of candidate LLMs on each query is available.
In contrast, ArenaExpert5K is used to assess whether each router can accurately predict the relative capability ranking of each model.

\paragraph{Dataset Preprocessing.}
For RouterBench, we sample 10,000 and 11 candidate models from the original dataset to construct our experimental subset.
For LLMRouterBench, we sample 12,732 queries and 11 candidate models.
The sampling is performed before the train-test split, so both splits are drawn from the same data distribution.
For ArenaExpert5K, we sample 2,678 queries and 105 candidate models.
Detailed preprocessing steps and candidate models are listed in Appendix \ref{app:expert5k-preprocess}. 

\paragraph{Data Split.}
Each dataset is partitioned into an 80\% training set $Q_t$ and a 20\% test set $Q$.
The training set $Q_t$ is used to build the Difficulty Analysis DB and to provide the retrieval pool or training data for the compared routers.
The test set $Q$ is used exclusively for evaluation.
All routers are trained and evaluated using the same split within each dataset to ensure a fair comparison.

\subsection{Metrics}
For RouterBench and LLMRouterBench, we follow \citet{song-etal-2025-irt} and \citet{feng2025graphrouter} and evaluate each routing method using Performance, Total Cost, and Reward, where Performance and Total Cost measure the answer correctness and total inference cost incurred by the selected model for each routing decision respectively.
Reward measures the performance-cost balance achieved by the selected model, as defined in Equation~\ref{eq:reward}.
We experiment in two $\alpha$ and $\beta$ settings to simulate different scenarios of cost-performance trade-off in real-world deployment.

For ArenaExpert5K, we use a different evaluation protocol since the dataset contains only pairwise human preference between models rather than the answer correctness for all models, making the reward metric incomputable.
A pairwise win of model $A$ over model $B$ only indicates that $A$ is preferred to $B$ for that specific comparison, and does not necessarily imply that $A$ is globally better across all queries or against all other candidate models.
In this scenario, we report Pairwise Accuracy, Spearman Correlation, and Total Cost.
Pairwise Accuracy measures whether the router correctly predicts the preferred model in a held-out pairwise comparison, and Spearman Correlation measures the consistency between the model ranking induced by the router and the Elo ranking estimated from the training comparisons.
Total Cost is computed based on the sum of the inference cost of each model selected by the router.
Details are in Appendix \ref{app:evaluation_metrics}.

\subsection{Difficulty Analyst LLMs}

To assess the VDAR-Router with different size of LLM as Difficulty Analyst, we use Gemma4-31B \citep{googledeepmind2026gemma4, farabet2026gemma4} for RouterBench and LLMRouterBench, and Qwen3.5-4B \citep{qwen35} for ArenaExpert5K.
To further evaluate the effect of smaller language model in real-world deployment, we also experiment with Qwen3.5-2B \citep{qwen35} on all datasets.
For embedding model, we use Qwen3-Embedding-0.6B \citep{qwen3embedding} for all datasets.

\subsection{Baselines}

We compare five routing methods in our experiments: KNN \citep{stripelis-etal-2024-tensoropera}, RouteLLM \citep{ong2025routellm}, RouterDC \citep{chen2024routerdc}, ICL-Router \citep{iclrouter}, and IRT-Router \citep{song-etal-2025-irt}. 
For KNN and RouterDC, we use the implementation provided by the \texttt{llmrouter} package~\citep{llmrouter2025}.
For RouteLLM, we select the introduced MLP classifier variant in our experiment.
For IRT-Router and ICL-Router, we implement based on the source code.
Specifically, we adopt Qwen3-embedding-0.6B for IRT-Router query embedding retrieval and use LoRA \citep{hu2022lora} to finetune Qwen3-4B \citep{qwen3technicalreport} for ICL-Router.
We also report two simple baselines on RouterBench and LLMRouterBench: always routing to the smallest LLM or the largest one (denoted as Small/Large LLM).
Details are in Appendix \ref{app:baseline-details}.

%% file: tables/01-bench-result.tex
\begin{table*}[t]
\centering
\resizebox{\textwidth}{!}{
\begin{tabular}{lrrrrrrrrrrrr}
\toprule

\multirow{3}{*}{\textbf{Router}} & \multicolumn{6}{c}{\textbf{RouterBench}} & \multicolumn{6}{c}{\textbf{LLMRouterBench}} \\
\cmidrule(lr){2-7} \cmidrule(lr){8-13}

& \multicolumn{3}{c}{\textbf{$\bm{\alpha = 0.8, \beta = 0.2}$}} & \multicolumn{3}{c}{\textbf{$\bm{\alpha = 0.6, \beta = 0.4}$}} 
& \multicolumn{3}{c}{\textbf{$\bm{\alpha = 0.8, \beta = 0.2}$}} & \multicolumn{3}{c}{\textbf{$\bm{\alpha = 0.6, \beta = 0.4}$}} \\ 
\cmidrule(lr){2-4} \cmidrule(lr){5-7} \cmidrule(lr){8-10} \cmidrule(lr){11-13} 

& \textbf{Perf.$\uparrow$} & \textbf{Cost$\downarrow$} & \textbf{Reward$\uparrow$} 
& \textbf{Perf.$\uparrow$} & \textbf{Cost$\downarrow$} & \textbf{Reward$\uparrow$} 
& \textbf{Perf.$\uparrow$} & \textbf{Cost$\downarrow$} & \textbf{Reward$\uparrow$} 
& \textbf{Perf.$\uparrow$} & \textbf{Cost$\downarrow$} & \textbf{Reward$\uparrow$} \\ 
\midrule
Small LLM & 30.52 & \textbf{0.09} & 24.15 & 30.52 & \textbf{0.09} & 17.77 & 41.67 & 23.12 & 30.93 & 41.67 & 23.12 & 20.19 \\ 
Large LLM &  77.20 & 6.67 & 41.76 & 77.20 & 6.67 &  6.32 & \underline{57.18} & 95.86 & 35.78 & \underline{57.18} & 95.86 & 14.38 \\
KNN \citep{stripelis-etal-2024-tensoropera} & 43.58 & \underline{0.15} & 34.40 & 43.58 & \underline{0.15} & 25.23 & 52.25 & 34.46 & 38.22 & 52.25 & 34.46 & 24.19 \\
RouterDC \citep{chen2024routerdc} & \underline{77.39} & 6.63 & 42.02 & \underline{77.39} & 6.63 & 6.65 & \underline{57.18} & 95.86 & 35.78 & \underline{57.18} & 95.86 & 14.38 \\
RouteLLM \citep{ong2025routellm} & 43.48 & \underline{0.15} & 34.34 & 43.48 & \underline{0.15} &  25.21 & 51.88 & \underline{3.79} & 41.11 & 51.88 & \underline{3.79} & 30.34 \\
IRT-Router \citep{song-etal-2025-irt} & 30.52 & \textbf{0.09} & 24.15 & 30.52 & \textbf{0.09} & 17.77 & 47.65 & \textbf{1.31} & 37.98 & 47.65 & \textbf{1.31} & 28.32 \\
ICL-Router \citep{iclrouter} & \textbf{77.44} & 6.60 & 42.15 & \textbf{77.44} & 6.60 & 6.85 & \textbf{60.69} & 94.86 & 38.69 & \textbf{60.69} & 94.86 & 16.69 \\
\hline
VDAR-Router (Gemma4-31B) & 75.02 & 3.80 & \textbf{48.61} &  74.14 & 2.52 & \underline{29.39} & 56.24 & 23.14 & \textbf{42.58} & 53.88 & 5.72 & \textbf{31.14} \\
VDAR-Router (Qwen3.5-2B) & 75.73 & 4.43 & \underline{47.31} &  73.32 & 2.38 & \textbf{29.71} & 55.59 & 19.88 & \underline{42.40} & 53.39 & 6.85 & \underline{30.61} \\
\bottomrule 
\end{tabular}
}
\caption{Results on RouterBench and LLMRouterBench dataset under different $\alpha$ and $\beta$ settings. Perf. stands for Performance, and Cost stands for Total Cost. The best result is in \textbf{boldface}, while the second best is \underline{underlined}.}
\label{tab:bench-result}
\end{table*}

%% file: sections/05-experiment-result.tex
\section{Experimental Results and Discussion}
Our experiments aim to answer the following research questions (RQs):

\noindent \textbf{RQ1:} How is the routing performance of VDAR-Router compared to baselines? (Section \ref{sec:quantitative_result})

\noindent \textbf{RQ2:} How does the hyper-parameter $k$ and $\alpha$ affect the routing performance of VDAR-Router? (Section \ref{sec:hyperparameter})

\noindent \textbf{RQ3:} What evidence is retrieved during the VDAR-Router routing? (Section \ref{sec:case_study})

\noindent \textbf{RQ4:} Are the retrieved top-$k$ queries really having a difficulty-level similar to the incoming query? (Section \ref{sec:rasch_ecdf})

\noindent \textbf{RQ5:} How much additional latency and cost does VDAR-Router incur during routing? (Section \ref{sec:routing-overhead}) 

\subsection{RQ1: Quantitative Results}
\label{sec:quantitative_result}
Table~\ref{tab:bench-result} summarizes the results on RouterBench and LLMRouterBench.
On RouterBench, VDAR-Router strikes the best performance-cost balance under both settings.
Although it does not obtain the highest raw performance, it substantially reduces inference cost compared to high-performance baselines such as ICL-Router and RouterDC, leading to better overall reward.
This suggests that the proposed method is not simply biased toward always selecting the strongest model, but can make more cost-effective routing decisions by incorporating reward ranking.
A similar trend is observed on LLMRouterBench.
VDAR-Router again achieves the best reward under both cost-performance preferences.
Compared with the RouteLLM and ICL-Router, which reach lowest cost and highest performance respectively, VDAR-Router better compromises performance and cost.
In addition, this advantage remains consistent when using a smaller 2B model as the Difficulty Analyst, suggesting that VDAR-Router can achieve strong routing performance across different analyst model sizes.
This indicates that difficulty-aware retrieval provides a useful routing signal beyond directly trained routing models or surface-level query similarity.

\input{tables/02-expert5k-result}

Table~\ref{tab:arena-expert5K-result} reports the results on ArenaExpert5K, where the target is to test whether each router can capture the nuance of model capability ranking while only pairwise human preference supervision is available.
VDAR-Router again achieves the best performance across all metrics.
Compared with the strongest baselines, VDAR-Router not only predicts pairwise preferences more accurately, but also produces a more reliable global ranking among candidate models.
This improvement also holds when replacing the Difficulty Analyst with Qwen3.5-2B, where VDAR-Router continues to outperform all baselines in both pairwise accuracy and global ranking correlation.
This indicates that VDAR-Router can better recognize the relative capability ranking of each models against incoming query.

\input{tables/03-train-test-split-expert5k}

\paragraph{Impact of Data Split Ratio.} To dive deeper into the efficiency and effectiveness of VDAR-Router, we further experiment with a 50\%:50\% train-test split ratio on each dataset to simulate a scenario with fewer training queries available.
As shown in Table~\ref{tab:train-test-split-experiment}, 
IRT-Router shows relatively competitive accuracy and correlation on ArenaExpert5K, indicating that adopting IRT and query embedding retrieval to estimate query difficulty can provide useful routing signals.
However, VDAR-Router still outperforms all baseline routers across all datasets, consistently demonstrating better performance-cost balance and producing a more reliable model capability ranking.
This result shows that difficulty analysis retrieval and reward ranking can more efficiently discern the capability difference between each model with fewer $Q_t$, thereby benefiting model routing.

\paragraph{Ablation Studies.} To understand the importance of each component in VDAR-Router, we conduct two ablation studies:
(1) removing the reward ranking and directly selecting the model with the highest estimated performance;
and (2) replacing the difficulty analysis retrieval with query retrieval.
The results are summarized in Table~\ref{tab:ablation-study}, and we can see that both variants lead to weaker routing performance compared with the full VDAR-Router.
When reward ranking is removed, the framework will only favor models with higher performance without explicitly considering inference cost.
Although this strategy preserves competitive raw performance, it loses the ability to balance performance improvement against additional cost, leading to weaker reward scores.
When the difficulty-analysis embedding is replaced by the raw query embedding, the router becomes more dependent on surface-level semantic similarity.
As a result, it retrieves queries that are textually similar but require different capabilities, or miss queries that are lexically different but share similar difficulty characteristics.
This shows that difficulty analysis provides a more suitable representation for retrieving examples that are relevant to model capability.

\input{tables/04-ablation-study}

\subsection{RQ2: Effects of Hyper-parameter}

\label{sec:hyperparameter}
\begin{figure}[t]
    \centering
    \begin{subfigure}[b]{\linewidth}
        \includegraphics[width=\linewidth]{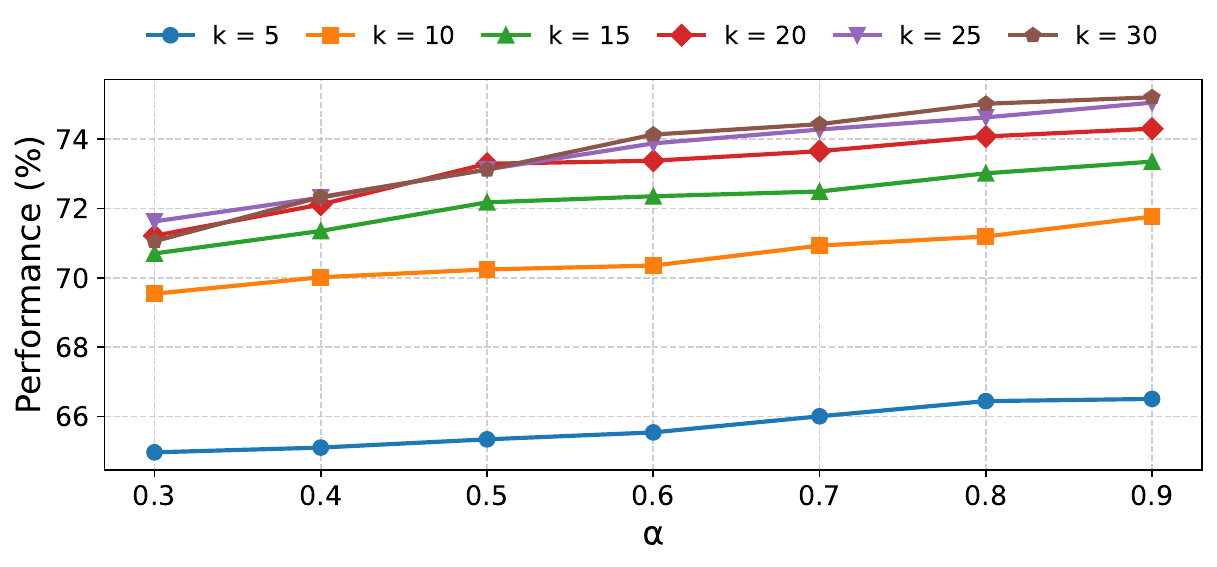}
        \caption{RouterBench}
    \end{subfigure}
    \hfill
    \begin{subfigure}[b]{\linewidth}
        \includegraphics[width=\linewidth]{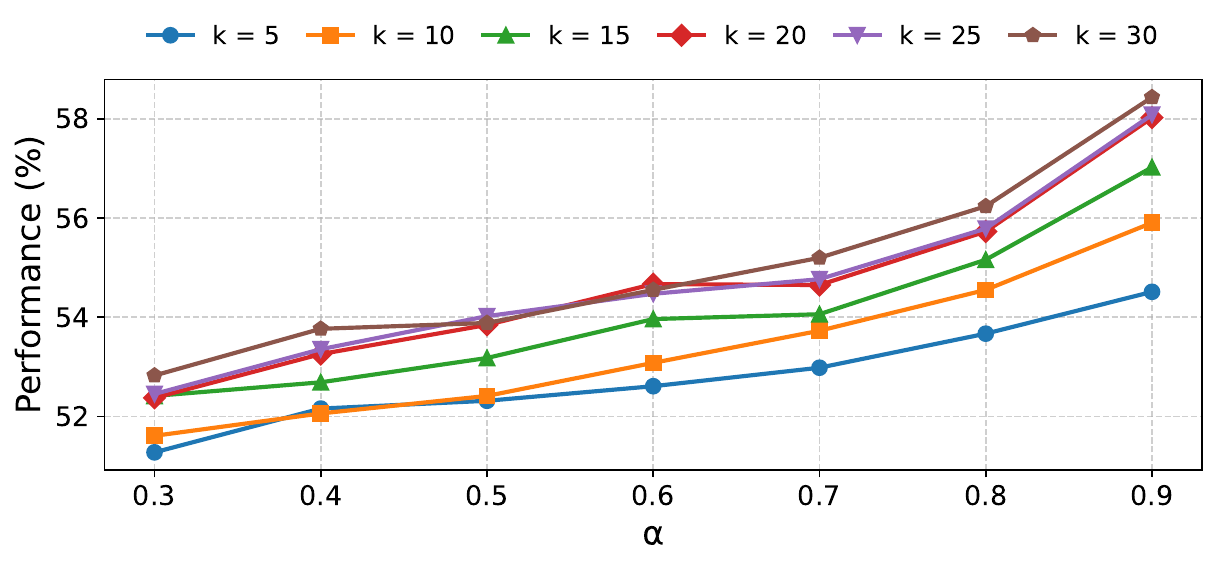}
        \caption{LLMRouterBench}
    \end{subfigure}
    \hfill
    \begin{subfigure}[b]{\linewidth}
        \includegraphics[width=\linewidth]{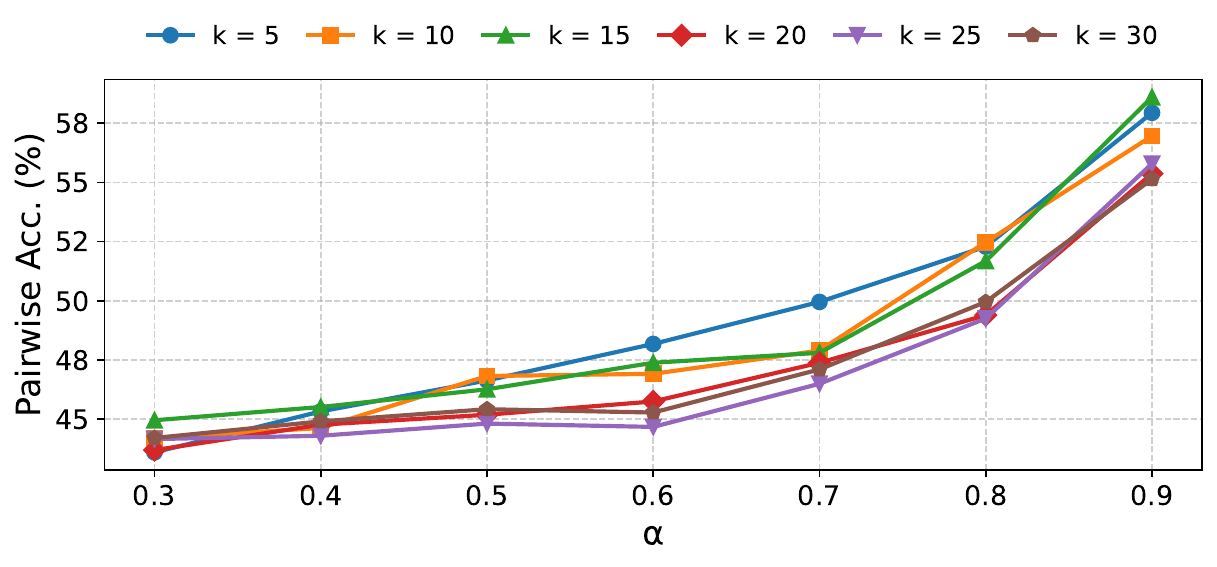}
        \caption{ArenaExpert5K}
    \end{subfigure}
    \caption{Routing performance under different $k$ and $\alpha$.}
    \label{fig:hyperparameter}
\end{figure}

We analyze the effect of the retrieval size $k$ and the reward weights $\alpha$ by setting $k \in \{5, 10, 15, 20, 25, 30\}$ and $\alpha \in \{0.3, 0.4, 0.5, 0.6, 0.7, 0.8, 0.9\}$, and the results are plotted in Figure~\ref{fig:hyperparameter}.
It can be observed that larger $\alpha$, which implies placing more emphasis on response quality during routing, generally leads to better performance and pairwise accuracy across all settings.
This is consistent with our design intuition that integrating the reward score during retrieval for model ranking can dynamically control the performance-cost balance.

We further analyze the effect of the retrieval size $k$. 
On RouterBench and LLMRouterBench, increasing $k$ generally yields stronger and more stable performance, suggesting that aggregating evidence from a broader set of difficulty-similar examples reduces the variance of model suitability estimation \citep{li-etal-2025-enhancing-retrieval}. 
This supports our intuition that queries with similar difficulty profiles tend to induce similar model capability rankings.
In contrast, the trend is less monotonic on ArenaExpert5K. 
We attribute this to its pairwise human preference supervision, where labels depend not only on query difficulty but also on response-relative factors such as style, verbosity, helpfulness, and subjective preference biases. 
Thus, enlarging the retrieval neighborhood may introduce examples that are difficulty-aligned but preference-heterogeneous, reducing the marginal gain from increasing $k$. 
This observation is in line with previous work on incorporating user personal preference \citep{dai2025personalizedrouter,tran2025arch} and points to a promising future direction of augmenting difficulty-aware retrieval with preference-aware or response-aware signals for subjective pairwise evaluation.

\subsection{RQ3: Case Study on Retrieval Targets}
\label{sec:case_study}

\begin{figure}[t]
    \input{figures/casestudy}
    \caption{Case study comparing raw query retrieval and difficulty-aware retrieval. The analysis has been truncated and summarized for better readability. $Ri$ denotes the $i$-th retrieved example ranked by similarity.}
    \label{fig:routing_case_study}
\end{figure}

To qualitatively uncover the difference between raw query retrieval and difficulty-aware retrieval, we conduct a case study on a sample from LLMRouterBench and depict the result in Figure~\ref{fig:routing_case_study}.\footnote{Full text of the case study is in Appendix \ref{app:full-case-study}.}
It can be seen that the standard KNN router retrieves examples that are superficially similar to the input query, sharing lexical and structural patterns such as mathematical notation, summations, binomial coefficients, and closed-form requests.
However, these retrieved queries do not necessarily require the same underlying reasoning capabilities.
Their similarity mainly reflects topical or notational overlap, which provides limited evidence for estimating which model is suitable for the target query.
In contrast, VDAR-Router retrieves examples using embeddings of difficulty analyses.
Specifically, the retrieved analyses emphasize similar capability requirements, such as recognizing hidden identities, applying polynomial or coefficient-based reasoning, performing combinatorial summation, and conducting symbolic manipulation.
These capability-aligned neighbors provide more informative evidence for estimating model performance on queries requiring similar reasoning skills.
As a result, VDAR-Router selects \texttt{glm-4.6}, one of the oracle models for this query.
This suggests that difficulty-analysis embeddings offer a more suitable retrieval space for routing, where neighborhood similarity better reflects model-relevant capability requirements rather than surface-form overlap.

\subsection{RQ4: Test-time Difficulty Alignment of Retrieved Queries}
\label{sec:rasch_ecdf}
\begin{figure}[t]
    \centering
    \begin{subfigure}[b]{.45\linewidth}
        \includegraphics[width=\linewidth]{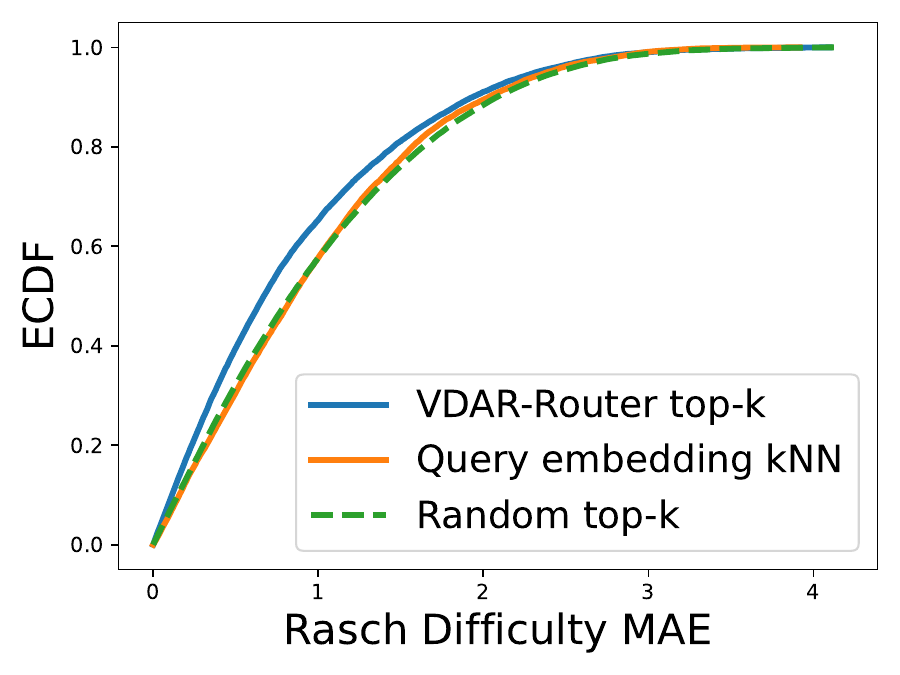}
        \caption{RouterBench}
    \end{subfigure}
    \hfill
    \begin{subfigure}[b]{.45\linewidth}
        \includegraphics[width=\linewidth]{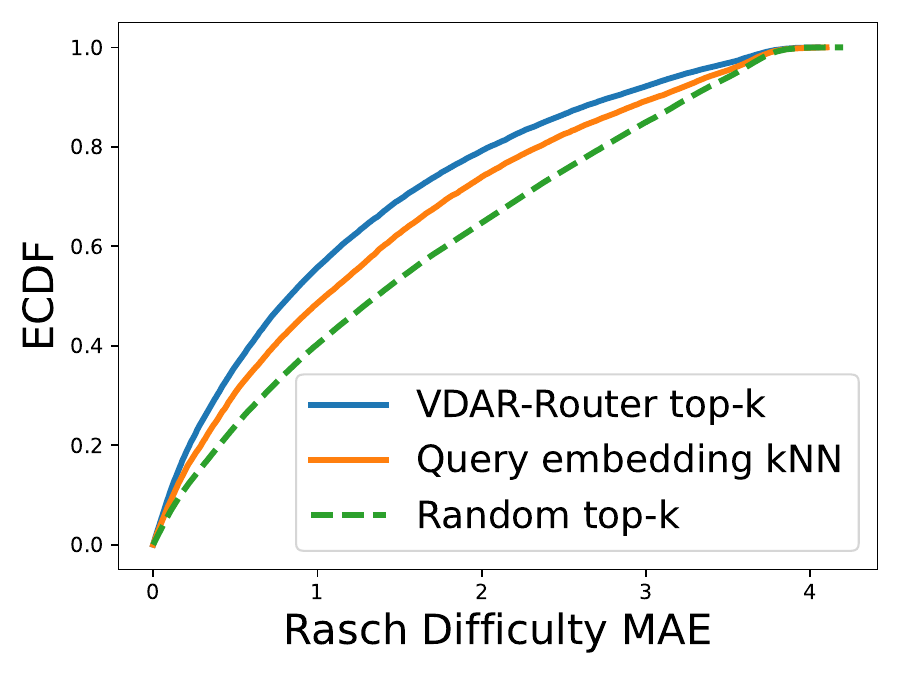}
        \caption{LLMRouterBench}
    \end{subfigure}
    \caption{Rasch difficulty difference between retrieved queries and incoming query.}
    \label{fig:rasch_ecdf}
\end{figure}

We quantitatively examine whether difficulty-aware retrieval actually retrieves examples with similar difficulty. 
For this analysis, we estimate numeric query difficulty using a Bayesian Rasch model \citep{rasch1993probabilistic,scarlatos-etal-2025-smart}, which is commonly used in IRT \citep{polo2024tinybenchmarks,fernandez2026radar} and Knowledge Tracing \citep{DisKT-2025} to model item difficulty.\footnote{Details of Rasch difficulty estimation are in Appendix \ref{app:rasch}.}
We measure the difference of Rasch difficulty between a test query $q$ and a retrieved training query $q'$ as $|b(q)-b(q')|$, where $b(\cdot)$ denotes the estimated Rasch difficulty, and a smaller value indicates better difficulty alignment.
We visualize the distribution of this difficulty gap using an empirical cumulative distribution function (ECDF).
In this plot, the $x$-axis represents the absolute Rasch difficulty gap, and the $y$-axis represents the proportion of retrieved samples whose difficulty gap is smaller than or equal to the corresponding $x$ value.
Therefore, a curve that rises earlier and stays closer to the upper-left corner indicates that more retrieved examples have similar difficulty.
We compare the proposed difficulty-analysis-based retrieval with two baselines: query-embedding retrieval, which follows the retrieval strategy used in IRT-Router as well as KNN Router, and random retrieval, which randomly samples training queries.

As shown in Figure \ref{fig:rasch_ecdf}, the proposed retrieval method retrieves examples with smaller difficulty gap than both baselines.
Compared with query-embedding retrieval, using difficulty analysis as the retrieval target results in retrieved examples that are more closely aligned with the test query in terms of estimated difficulty.
Compared with random retrieval, the improvement further confirms that the observed alignment is not simply due to chance.
This consolidates the design motivation of VDAR-Router that historical examples are more useful for routing when they reflect similar difficulty characteristics to the incoming query.

\subsection{RQ5: Routing Overhead Analysis}
\label{sec:routing-overhead}

\begin{table*}[t]
\centering
\resizebox{0.9\linewidth}{!}{
\begin{tabular}{lccccc}
\toprule
\textbf{Metric}
& \textbf{Difficulty Analysis}
& \textbf{Embedding}
& \textbf{FAISS Retrieval}
& \textbf{Reward Ranking}
& \textbf{End-to-End} \\
\midrule
Mean latency/query
& 1.4300
& 0.0210
& 0.0040
& 0.0001
& 1.4600 \\
\bottomrule
\end{tabular}
}
\caption{
Mean routing latency (seconds) per query using Qwen3.5-2B as the Difficulty Analyst.
}
\label{tab:latency}
\end{table*}

VDAR-Router introduces an auxiliary Difficulty Analyst before retrieval and model selection.
To account for the complete routing overhead, we measure the latency of difficulty analysis generation, embedding, retrieval, and reward ranking using Qwen3.5-2B as the Difficulty Analyst, and the results are summarized in Table~\ref{tab:latency}.
The latency breakdown shows that difficulty-analysis generation accounts for most of the routing overhead, while embedding, retrieval, and reward ranking introduce only minor additional latency.
The complete routing pipeline requires approximately 1.46 seconds per query on average, remaining in acceptable bounds for real-world applications.

We further incorporate the input and output token cost of the Difficulty Analyst into the routing cost term using the same OpenRouter pricing protocol as the main experiments.
Table~\ref{tab:online-cost} reports the resulting cost-adjusted performance.
After accounting for the auxiliary analyst cost, VDAR-Router remains above the strongest baseline across all four RouterBench and LLMRouterBench reward settings.
A lightweight analyst therefore provides a practical configuration while preserving the performance--cost advantage of VDAR-Router.

\begin{table*}[t]
\centering
\resizebox{\linewidth}{!}{
\begin{tabular}{lcccccc}
\toprule
\textbf{Router}
& \textbf{RB $\alpha=0.8$}
& \textbf{RB $\alpha=0.6$}
& \textbf{LRB $\alpha=0.8$}
& \textbf{LRB $\alpha=0.6$}
& \textbf{Expert5K Acc.}
& \textbf{Expert5K Cost} \\
\midrule
KNN
& 34.40 & 25.23 & 38.22 & 24.19 & 47.71 & 0.0405 \\
RouterDC
& 42.02 & 6.65 & 35.78 & 14.38 & 48.08 & 0.0194 \\
RouteLLM
& 34.34 & 25.21 & 41.11 & 30.34 & 52.85 & 0.0365 \\
IRT-Router
& 24.15 & 17.77 & 37.98 & 28.32 & 53.22 & 0.0016 \\
ICL-Router
& 42.15 & 6.85 & 38.69 & 16.69 & 49.11 & 0.0349 \\
\midrule
VDAR-Router (Qwen3.5-2B)
& \textbf{48.32}
& \textbf{29.36}
& \textbf{43.02}
& \textbf{31.28}
& \textbf{58.90}
& \textbf{0.0005} \\
\bottomrule
\end{tabular}
}
\caption{
Routing performance after incorporating the input and output token cost of the Qwen3.5-2B Difficulty Analyst into VDAR-Router.
}
\label{tab:online-cost}
\end{table*}

%% file: tables/02-expert5k-result.tex
\begin{table}[t]
\centering
\resizebox{\linewidth}{!}{
\begin{tabular}{lrrr}
\toprule
\textbf{Router} & \multicolumn{1}{c}{\textbf{Acc.}} & \multicolumn{1}{c}{\textbf{SP}} & \multicolumn{1}{c}{\textbf{Cost}} \\
\midrule
KNN   & 47.71 & 0.0157 & 0.0405 \\
RouterDC & 48.08 & -0.0875 & 0.0194 \\
RouteLLM   & 52.85 & 0.0710 & 0.0365 \\
IRT-Router & 53.22 & 0.1851 & 0.0016 \\
ICL-Router & 49.11 & -0.0967 & 0.0349 \\
\hline
VDAR-Router (Qwen3.5-4B) & \underline{57.66} & \underline{0.5099} & \underline{0.0006} \\
VDAR-Router (Qwen3.5-2B) & \textbf{60.28} & \textbf{0.5137} & \textbf{0.0004} \\
\bottomrule
\end{tabular}%
}
\caption{Results on ArenaExpert5K dataset. Acc. stands for Pairwise Accuracy, SP stands for Spearman Correlation, and Cost stands for Total Cost.}
\label{tab:arena-expert5K-result}
\end{table}

%% file: tables/03-train-test-split-expert5k.tex
\begin{table}[t]
\centering
\resizebox{\linewidth}{!}{
\begin{tabular}{lcccc}
\toprule
\multirow{2}{*}{\textbf{Router}} & \multicolumn{1}{c}{\textbf{RouterBench}} & \multicolumn{1}{c}{\textbf{LLMRouterBench}} & \multicolumn{2}{c}{\textbf{ArenaExpert5K}} \\
\cmidrule(lr){2-2} \cmidrule(lr){3-3} \cmidrule(lr){4-5}
& \textbf{Reward} & \textbf{Reward}
& \textbf{Acc.} & \textbf{SP} \\ 
\midrule
KNN   & \underline{44.43} & 42.06 & 47.35 & 0.0190 \\
RouterDC & 33.98 & 42.30 & 47.71 & 0.0884 \\
RouteLLM   & 33.98 & \underline{42.32} & 50.32 & 0.1346 \\
IRT-Router & 23.80 & 39.76 & \underline{53.09} & \underline{0.1626} \\
ICL-Router & 44.06 & 38.83 & 47.20 &  -0.0493 \\
\hline
VDAR-Router & \textbf{53.73} & \textbf{42.93} & \textbf{58.16} & \textbf{0.4514} \\
\bottomrule 
\end{tabular}
}
\caption{Results of 50\%:50\% train-test split ratio. Reward metric is calculated with $\alpha=0.8$ and $\beta=0.2$.}
\label{tab:train-test-split-experiment}
\end{table}

%% file: tables/04-ablation-study.tex
\begin{table}[t]
\centering
\resizebox{\linewidth}{!}{
\begin{tabular}{lcccc}
\toprule
\multirow{2}{*}{\textbf{Router}} & \multicolumn{1}{c}{\textbf{RouterBench}} & \multicolumn{1}{c}{\textbf{LLMRouterBench}} & \multicolumn{2}{c}{\textbf{ArenaExpert5K}} \\
\cmidrule(lr){2-2} \cmidrule(lr){3-3} \cmidrule(lr){4-5}
& \textbf{Reward} & \textbf{Reward}
& \textbf{Acc.} & \textbf{SP} \\ 
\midrule
VDAR-Router & \textbf{48.61} & \textbf{42.58} & \textbf{57.66} & \textbf{0.5099} \\
\hline
w/o Reward Reranking  & 45.97 & 38.11 & 49.44 & 0.0129 \\
w/o Difficulty Analysis Retrieval  & 42.39 & 39.96 & 54.63 & 0.5093 \\
\bottomrule 
\end{tabular}
}
\caption{Results of ablative experiments. Reward metric is calculated with $\alpha=0.8$ and $\beta=0.2$.}
\label{tab:ablation-study}
\end{table}

%% file: figures/casestudy.tex
\newtcolorbox{casebox}{
  colback=white,
  colframe=black,
  boxrule=0.6pt,
  arc=0pt,
  left=4pt,
  right=4pt,
  top=4pt,
  bottom=4pt,
  before skip=6pt,
  after skip=6pt
}

\newcommand{\caseseparator}{
  \par\vspace{4pt}
  \noindent\hrule height 0.5pt
  \vspace{4pt}
}

\definecolor{matchIdentity}{HTML}{D55E00}      
\definecolor{matchPoly}{HTML}{7B2CBF}          
\definecolor{matchComb}{HTML}{0072B2}          
\definecolor{matchSymbolic}{HTML}{009E73}      
\definecolor{matchRisk}{HTML}{CC79A7}          

\newcommand{\idmatch}[1]{\textcolor{matchIdentity}{#1}}
\newcommand{\polymatch}[1]{\textcolor{matchPoly}{#1}}
\newcommand{\combmatch}[1]{\textcolor{matchComb}{#1}}
\newcommand{\symmatch}[1]{\textcolor{matchSymbolic}{#1}}
\newcommand{\riskmatch}[1]{\textcolor{matchRisk}{#1}}

\newcommand{\bad}[1]{\textcolor{red}{#1}}
\newcommand{\good}[1]{\textcolor{blue}{#1}}
\newcommand{\cmark}{\ding{51}} 
\newcommand{\xmark}{\ding{55}} 

\centering
\footnotesize
\begin{casebox}

\colorbox{gray!20}{\textbf{Query:}}
\textbf{Find a closed form for}
\[
\textstyle
\sum_{k=0}^{n}
\left((2k+1)^5
\binom{2k}{k}
\binom{2n-2k}{n-k}\right).
\]

\textbf{Query Difficulty:}
\combmatch{weighted binomial convolution},
\combmatch{central binomial coefficients},
\idmatch{convolution identities},
\polymatch{generating functions},
\polymatch{polynomial decomposition},
\symmatch{symbolic manipulation},
\riskmatch{algebraic drift}.

\caseseparator

\textbf{KNN Router}
\hfill
\textit{qwen3-235b-a22b-2507 (\xmark)}

\textbf{R1:}
Find the sum of all integers $k$ such that
$\binom{23}{4}+\binom{23}{5}=\binom{24}{k}$.

\textbf{R2:}
Find the function $f$ with the lowest complexity such that
$\textstyle \sum_{m=0}^{n}(-1)^m {n\choose m}
\frac{\Gamma(\frac{3}{2}+n)}
{\Gamma(\frac{3}{2}+n-m)}
\leq Cf(n)$
for all $n\in\mathbb{N}$.

\textbf{R3:}
Find a closed expression of the infinite product
$\textstyle \prod_{n=0}^{\infty}(1-e^{-(2n+1)\pi})$.

\caseseparator

\textbf{VDAR Router}
\hfill
\textit{glm-4.6 (\cmark)}

\textbf{R1:}
\symmatch{multi-step mathematical insight},
\idmatch{recognition of hidden identity},
\polymatch{polynomial coefficient reasoning},
\combmatch{combinatorial counting}.

\textbf{R2:}
\combmatch{summation-order manipulation},
\combmatch{counting repeated terms},
\combmatch{discrete combinatorial reasoning},
\symmatch{symbolic derivation}.

\textbf{R3:}
\idmatch{specialized combinatorial identities},
\combmatch{binomial symmetry},
\symmatch{symbolic reasoning},
constraint verification.

\end{casebox}

%% file: sections/06-conclusion.tex
\section{Conclusion}

In this work, we introduce VDAR-Router, a difficulty-aware retrieval-based framework for LLM routing.
VDAR-Router analyzes the difficulty characteristics and the required capability for a correct model response of an input query, retrieves historically observed examples with similar difficulty patterns, and ranks candidate model suitability based on their performance and cost of the retrieval results.
This design facilitates a training-free and more human-interpretable routing decision, while reducing the dependence on explicit routing labels, complete prompt-level scores, or additional router training.
VDAR-Router consistently achieves the strongest cost-performance trade-off on three datasets, showcasing the effectiveness of our framework.
We believe our results can serve as a stepping stone for more research on adaptive and difficulty-aware model selection, and several future directions could be further explored under VDAR-Router framework, such as structured or latent difficulty representations and stronger integration with sparse preference-based supervision.

%% file: sections/98-limitation.tex
\section*{Limitations}

While our proposed framework demonstrates strong results, several limitations remain, pointing to directions for future work.

\paragraph{Test-time difficulty analysis generation.}
We note that the success of the entire framework highly depends on the generation and retrieval accuracy of the difficulty analysis, which may introduce additional latency.
Although we have demonstrated that adopting smaller model can still reach better routing performance than all baseline methods, future research will be further explored to mitigate this test-time inference cost, such as finetuning a small language model to generate latent difficulty analysis tokens.

\paragraph{Difficulty similarity and preference heterogeneity.}
We assume that queries with similar difficulty profiles tend to induce similar model capability rankings.
However, this assumption may not always hold in preference-based settings, where labels can depend not only on task difficulty but also on response style, verbosity, formatting, helpfulness, and subjective preference biases.
This may explain why increasing the retrieval size does not always lead to monotonic improvement on ArenaExpert5K.
Future work could combine difficulty-aware retrieval with preference-aware or response-aware signals to better handle subjective pairwise evaluation scenarios.

%% file: sections/99-appendix.tex
\section*{Appendix}

\section{Difficulty Analysis Prompt}
\label{app:difficulty-analysis-prompt}
The system prompt we use for the Difficulty Analyst is shown in Figure \ref{fig:difficulty-analysis-prompt}.

\newtcolorbox{promptbox}{
    breakable,
    colback=gray!3,
    colframe=gray!45,
    boxrule=0.4pt,
    arc=2pt,
    left=8pt,
    right=8pt,
    top=8pt,
    bottom=8pt,
    fontupper=\ttfamily\footnotesize\raggedright,
}

\begin{promptbox}
Your role as an assistant is to analyze the difficulty of a given query for a large language model
through a systematic long thinking process analysis. You will be provided with the user query.
You need to evaluate the incoming query on several key dimensions: reasoning, comprehension,
instruction following, agentic, knowledge retrieval, coding, multilingual. For each dimension,
elaborate on the specific challenges and required capabilities. Please structure your response into Summary.
In the Summary section, based on the analysis, explorations, and reflections from the Think section, systematically present the summary you think for the query difficulty.
The summary should remain a clear, concise expression style and detail necessary difficulty description to reach the conclusion, formatted as follows: <summary> {final formatted, precise, and clear summary} </summary>.
Now, try to analyze the following query through the above guidelines:
\end{promptbox}
\captionof{figure}{Prompt used for query difficulty analysis.}
\label{fig:difficulty-analysis-prompt}

\section{ArenaExpert5K Evaluation Details}
\label{app:expert5k-evaluation}

\subsection{Dataset Preprocess}
\label{app:expert5k-preprocess}

We first normalize each model name into its corresponding OpenRouter\footnote{\url{https://openrouter.ai/}} model identifier and retrieve its per-token API price.
If a model name cannot be converted into a valid OpenRouter identifier, or if its pricing information is unavailable, we use the price of \texttt{openai/gpt-4o-mini} as the fallback price.

In addition, we remove records labeled as \textit{both bad} or \textit{tie}, since these cases do not provide a clear preference signal for comparing the two candidate answers.
For each remaining pairwise comparison, we convert the pair into two model-level records: the winning model is assigned a reward of 1, while the losing model is assigned a reward of 0.
To reduce potential distribution imbalance, we also ensure that every candidate model appears at least once in both the training and test splits.

We further filter out prompt pairs with empty answers and those whose \texttt{evaluation\_order} is not equal to 1, retaining only meaningful first-order preference records for training and testing.
After preprocessing, the final ArenaExpert5K subset contains 2,678 queries and 5,500 answers across 105 models.
The complete candidate models of each dataset are summarized in Table~\ref{tab:model_list}.
\subsection{Evaluation Metrics}
\label{app:evaluation_metrics}
\paragraph{Pairwise Accuracy.}

Each test instance contains a query $q_i$, two candidate models, and a preference label.
Let $m_i^{+}$ be the preferred model and $m_i^{-}$ be the non-preferred model.
Pairwise accuracy is computed as:
\begin{equation}
\small
    \mathrm{PairwiseAcc}
    =
    \frac{1}{N}
    \sum_{i=1}^{N}
    \mathbb{I}
    \left[
        R(m_i^{+} \mid q_i)
        >
        R(m_i^{-} \mid q_i)
    \right],
\end{equation}
where $R(m \mid q_i)$ is the router reward score.

\paragraph{Spearman Correlation.}

For each query $q_i$, the router produces a model ranking $\pi_i^{R}$ according to the reward scores.
We compare this ranking with the reference ranking $\pi^{\mathrm{Elo}}_{\mathrm{train}}$, which is constructed from the preference records in the training split using Elo rating.
The Spearman correlation for query $q_i$ is computed as:
\begin{equation}
    \rho_i
    =
    \mathrm{Spearman}
    \left(
        \pi_i^{R},
        \pi^{\mathrm{Elo}}_{\mathrm{train}}
    \right).
\end{equation}

The final score is averaged over all test queries:
\begin{equation}
    \mathrm{SpearmanCorr}
    =
    \frac{1}{N}
    \sum_{i=1}^{N}
    \rho_i.
\end{equation}

\paragraph{Evaluation Cost.}

During evaluation, we report the total routing cost of the top-1 selected models.
Since the output length of the selected model is unavailable, we only compute the input token cost.
This simulates the observable API cost incurred when the router sends each query to its predicted top-1 model.

Let
\begin{equation}
    \hat{m}_i
    =
    \arg\max_{m \in M}
    R(m \mid q_i)
\end{equation}
be the model selected by the router for query $q_i$.
The total evaluation cost is computed as:
\begin{equation}
    \mathrm{TotalCost}
    =
    \sum_{i=1}^{N}
    n_{\mathrm{in}}(q_i)
    \cdot
    p_{\mathrm{in}}(\hat{m}_i),
\end{equation}
where $n_{\mathrm{in}}(q_i)$ is the input token length of query $q_i$, and $p_{\mathrm{in}}(\hat{m}_i)$ is the input token price of the selected model.

\subsection{Runtime Cost Estimation}

During routing, the router estimates the expected cost of each candidate model using both input and output token costs.
Since the actual output length is unavailable before generation, we approximate it using the average output length of the retrieved examples.

For query $q$ and candidate model $m$, the estimated runtime cost is:
\begin{equation}
\small
    \widehat{\mathrm{Cost}}(q,m)
    =
    n_{\mathrm{in}}(q) \cdot p_{\mathrm{in}}(m)
    +
    \widehat{n}_{\mathrm{out}}(q,m) \cdot p_{\mathrm{out}}(m),
\end{equation}
where $n_{\mathrm{in}}(q)$ is the input token length, $p_{\mathrm{in}}(m)$ and $p_{\mathrm{out}}(m)$ are the input and output token prices, and $\widehat{n}_{\mathrm{out}}(q,m)$ is the estimated output length.

The estimated output length is computed as:
\begin{equation}
    \widehat{n}_{\mathrm{out}}(q,m)
    =
    \frac{1}{|\mathcal{I}_m(q)|}
    \sum_{i \in \mathcal{I}_m(q)}
    n_{\mathrm{out}}(q_i,m),
\end{equation}
where $\mathcal{I}_m(q)$ denotes the retrieved examples associated with model $m$.

The estimated runtime cost $\widehat{\mathrm{Cost}}(q,m)$ is then used as the cost term in the original reward function (Equation \ref{eq:reward}) to compute the final routing score for each candidate model.

\section{Baseline Details}
\label{app:baseline-details}

\subsection{KNN}

KNN is a retrieval-based router that performs model selection based on similar queries in the training set.
For each input query, it retrieves the top-$k$ nearest training examples using query embeddings.
The router then aggregates the model performance or preference signals from the retrieved examples to estimate the suitability of each candidate model.
In our experiments, KNN uses the original query embedding for retrieval.
This differs from VDAR-Router, which performs retrieval over difficulty analysis representations rather than the raw query representation.

\subsection{RouteLLM}

RouteLLM is a supervised routing framework for cost-performance-aware model selection.
Among the router variants discussed in RouteLLM, we adopt the MLP-based router in our experiments.
The router takes the query representation as input and predicts a score over candidate models.
During inference, the candidate model with the highest predicted score is selected as the routing result.

\subsection{RouterDC}

RouterDC is a learned router that models the compatibility between queries and candidate models.
It learns query-side and model-side representations and predicts how suitable each model is for a given input query.
During inference, RouterDC computes a compatibility score for each candidate model and selects the model with the highest score.

\subsection{IRT-Router}

IRT-Router predicts the performance of each candidate LLM on a given query using an item-response-theory-based model.
Following the original formulation, it represents both queries and LLMs as embeddings, where the LLM embedding is constructed from the model profile.
The router then estimates the interaction between query attributes and LLM abilities to predict model performance.

The original paper proposes two variants, MIRT-Router and NIRT-Router.
MIRT-Router uses a multidimensional IRT formulation, while NIRT-Router further incorporates a relevance vector to associate queries with ability dimensions.
The predicted performance is finally combined with fixed model cost to determine the routed model.
In our experiments, we employ the MIRT-Router variant.

\subsection{ICL-Router}

ICL-Router is a trainable routing method that represents candidate models through in-context learned capability profiles.
It uses an embedding model and a projector to convert queries into vector representations, which are then interpreted by an LLM-based router.
The method is trained in two stages: query reconstruction training aligns the projected query vectors with the router's semantic space, while model routing training teaches the router to predict whether a candidate model can correctly answer a query.
For each candidate model, ICL-Router constructs a capability profile from representative query-performance pairs.
At inference time, the router combines the input query vector with each model's capability profile and predicts the probability that the model will answer correctly.
The model with the highest predicted probability is selected.

\section{Rasch Difficulty Estimation Implementation}
\label{app:rasch}
We estimate query-level difficulty with a Bayesian Rasch model, treating each query as an item and each candidate LLM as an examinee.
Given the observed correctness score $x_{mq}\in[0,1]$ of model $m$ on query $q$, we model the response probability as $p(x_{mq}=1 \mid \theta_m,b(q)) = \sigma(\theta_m-b(q))$, where $\theta_m$ denotes the latent ability of model $m$ and $b(q)$ denotes the latent difficulty of query $q$.
Since our routing labels may be fractional success rates rather than binary outcomes, we optimize the likelihood using binary cross-entropy with continuous targets.
We place independent zero-mean Gaussian priors on both model abilities and query difficulties, and approximate the posterior with a mean-field Gaussian variational distribution.
The variational parameters are optimized by minimizing the stochastic negative ELBO, which consists of a minibatch-scaled binary cross-entropy term and closed-form KL regularization against the Gaussian priors.

To place training and test queries on a common Rasch scale, we use a two-stage estimation procedure.
First, we jointly infer model abilities and training-query difficulties from the training routing responses.
Second, we freeze the learned posterior mean of each model ability and infer only the difficulties of test queries using the test routing responses.
This anchoring step ensures that both training and test query difficulties are measured relative to the same set of model abilities, making them comparable for our downstream proximity analysis.
We use the posterior mean of $b(q)$ as the final Rasch difficulty score for each query, where larger values indicate harder queries.
Algorithm~\ref{alg:rasch} summarizes the fitting procedure.

\begin{table*}[t]
\centering
\resizebox{\linewidth}{!}{
\begin{tabular}{lcccccc}
\toprule
\textbf{Representation / Prompt}
& \textbf{RB $\alpha=0.8$}
& \textbf{RB $\alpha=0.6$}
& \textbf{LRB $\alpha=0.8$}
& \textbf{LRB $\alpha=0.6$}
& \textbf{Expert5K Acc.}
& \textbf{Expert5K SP.} \\
\midrule
Strongest baseline
& 42.15 & 25.23 & 41.11 & 30.34 & 53.22 & 0.1851 \\
Raw query embedding
& 42.39 & -- & 39.96 & -- & 54.63 & 0.5093 \\
\midrule
3-dimension difficulty
& 48.56 & 29.93 & 42.59 & 30.81 & 61.07 & 0.5356 \\
7-dimension difficulty, concise
& 48.63 & 29.98 & 43.06 & 31.35 & 58.90 & 0.5367 \\
12-dimension difficulty
& 49.21 & \textbf{30.45} & 42.54 & \textbf{31.37} & 56.96 & \textbf{0.5427} \\
Free-text difficulty, no taxonomy
& \textbf{49.27} & 30.36 & \textbf{43.14} & 31.07 & \textbf{61.92} & 0.5370 \\
\bottomrule
\end{tabular}
}
\caption{
Effect of different difficulty-analysis representations.
}
\label{tab:appendix-prompt-ablation}
\end{table*}

\section{Robustness to Difficulty Representation}
\label{app:difficulty-representation}

The default VDAR-Router uses seven interpretable dimensions to describe query difficulty and required model capabilities.
To examine whether the routing improvement depends on this particular taxonomy, we conduct controlled experiments by varying only the Difficulty Analyst prompt while fixing the Qwen3.5-2B analyst, embedding model, retrieval size, and reward formulation.
We compare a coarse 3-dimension representation, the default 7-dimension representation, a fine-grained 12-dimension representation, and a free-text difficulty analysis without an explicit taxonomy.
All four difficulty-oriented representations outperform the strongest baseline across the evaluated RouterBench and LLMRouterBench settings, while no single number of dimensions is uniformly optimal.
These results show that VDAR-Router does not rely on the handcrafted seven-dimension taxonomy.
Instead, the taxonomy primarily provides an interpretable structure for describing query difficulty and required capabilities.
The consistent advantage over raw-query retrieval further indicates that difficulty-oriented representations provide more useful routing evidence than semantic similarity alone.

\section{Use of AI assistants}
We use AI to help our coding and grammar correction in our paper writing.
We do not use AI to help us in related paper surveys.

\clearpage
\onecolumn

\section{Full Case Study} \label{app:full-case-study}
\input{figures/appendix-casestudy-full}

\clearpage
\twocolumn

\input{tables/06-modellist}

\lstset{
  language=Python,
  basicstyle=\small\ttfamily,
  keywordstyle=\bfseries,
  commentstyle=\itshape\color{gray!70!black},
  showstringspaces=false,
  columns=fullflexible,
  xleftmargin=1em,
}
\begin{algorithm*}[t]
\caption{Bayesian Rasch difficulty estimation. Stage~1 jointly fits model abilities and training-query difficulties. Stage~2 fixes the learned model-ability means and optimizes only test-query difficulties on test responses.}
\label{alg:rasch}
\begin{lstlisting}
# Inputs:
#   query_ids, model_ids, labels  : tensors of shape [N], labels in [0,1]
#   sigma_theta, sigma_beta       : prior standard deviations
#   B, T, lr, eps                 : batch size, steps, learning rate, SD floor

# --- Warm-start variational means from marginal success rates ---
mu_theta = center(logit(model_success_rate))    # shape [M]
mu_beta  = center(-logit(query_success_rate))   # shape [Q]
rho_theta = full([M], softplus_inv(0.1))        # SD via s = softplus(rho) + eps
rho_beta  = full([Q], softplus_inv(0.1))

params = [mu_theta, rho_theta, mu_beta, rho_beta]
opt    = Adam(params, lr=lr)

for step in range(T):
    # --- Mini-batch of response triples ---
    idx = randint(0, N, size=B)
    qi, mi, yi = query_ids[idx], model_ids[idx], labels[idx]

    s_theta = softplus(rho_theta) + eps         # [M]
    s_beta  = softplus(rho_beta)  + eps         # [Q]

    # --- Reparameterised samples for the batch ---
    theta = mu_theta[mi] + s_theta[mi] * randn(B)
    beta  = mu_beta[qi]  + s_beta[qi]  * randn(B)

    # --- Fractional BCE log-likelihood over the batch ---
    nll = bce_with_logits(theta - beta, yi, reduction='sum')

    # --- Closed-form KL( N(mu, s^2) || N(0, sigma^2) ), summed over dims ---
    kl_theta = gaussian_kl(mu_theta, s_theta, sigma_theta)
    kl_beta  = gaussian_kl(mu_beta,  s_beta,  sigma_beta)

    # --- Stochastic ELBO with full-data likelihood rescaling ---
    loss = (N / B) * nll + kl_theta + kl_beta

    opt.zero_grad(); loss.backward()
    clip_grad_norm_(params, max_norm=10.0)
    opt.step()

return mu_theta, softplus(rho_theta) + eps, mu_beta, softplus(rho_beta) + eps
\end{lstlisting}
\end{algorithm*}

%% file: figures/appendix-casestudy-full.tex
\tcbset{
  casebox/.style={
    enhanced,
    breakable,
    colback=white,
    colframe=black!40,
    colbacktitle=black!45,
    coltitle=white,
    boxrule=0.5pt,
    arc=1mm,
    left=1.4mm,
    right=1.4mm,
    top=0.8mm,
    bottom=0.8mm,
    fonttitle=\bfseries\footnotesize,
    fontupper=\footnotesize,
  }
}

\setlist[enumerate]{leftmargin=*, itemsep=2pt, topsep=2pt}

\noindent
The label ``Rank $n$'' corresponds to $R_n$ in Figure~\ref{fig:routing_case_study}, denoting the $n$-th retrieved example ranked by retrieval similarity.

\subsection{Question}
\label{appendix:case_study_full}

\begin{tcolorbox}[casebox, title=Input Question]
Find a closed form for the following sum:
$\sum_{k=0}^{n}((2k+1)^5\binom{2k}{k}\binom{2n-2k}{n-k})$
\end{tcolorbox}

\begin{tcolorbox}[casebox, title=Difficulty Analysis]
based on the provided guidelines and the context of previous analyses, here is the systematic difficulty analysis for the query: **"find a closed form for the following sum: $\sum_{k=0}^{n}((2k+1)^5\binom{2k}{k}\binom{2n-2k}{n-k})$"**

\#\#\# **overall difficulty assessment: high**

this query is a sophisticated mathematical challenge that moves beyond simple pattern matching or basic formula application. it requires a synthesis of advanced combinatorial identities, symbolic manipulation, and rigorous verification.

---

\#\#\# **dimension-by-dimension analysis**

\#\#\#\# **1. knowledge retrieval (high difficulty)**
the model cannot rely on a simple lookup table for this specific sum. it requires retrieval of several specialized mathematical tools:
*   **central binomial coefficients:** the model must recognize the properties of $\binom{2k}{k}$ and $\binom{2n-2k}{n-k}$.
*   **convolution identities:** the core of the problem relies on the identity $\sum_{k=0}^n \binom{2k}{k}\binom{2n-2k}{n-k} = 4^n$. the model must retrieve this specific identity as the "base" for the more complex sum.
*   **generating functions:** to handle the $(2k+1)^5$ term, the model may need to retrieve knowledge of generating functions for central binomial coefficients, specifically the identity $\frac{1}{\sqrt{1-4x}} = \sum_{n=0}^\infty \binom{2n}{n} x^n$.

\#\#\#\# **2. reasoning (very high difficulty)**
the primary challenge is the polynomial term $(2k+1)^5$. this requires a multi-step reasoning process:
*   **polynomial decomposition:** the model must decide how to handle $(2k+1)^5$. it cannot simply distribute the power; it must express $(2k+1)^5$ in terms of falling factorials or operators (like $x \frac{d}{dx}$) that interact cleanly with the binomial coefficients.
*   **operator application:** if using generating functions, the model must apply the operator $(2x \frac{d}{dx} + 1)^5$ to the generating function of the convolution. this involves repeated differentiation and algebraic simplification, where the risk of a "calculation hallucination" is extremely high.
*   **symmetry exploitation:** the model must recognize the symmetry between $k$ and $n-k$ to simplify the sum, potentially splitting the polynomial into symmetric and anti-symmetric parts to reduce the number of terms to calculate.

\#\#\#\# **3. comprehension (moderate difficulty)**
*   **latex parsing:** the model must correctly parse the latex notation, specifically the exponents and the binomial coefficients.
*   **goal identification:** the term "closed form" implies that the final answer must be a function of $n$ without any summation symbols ($\sum$), which is a strict constraint on the output format.

\#\#\#\# **4. instruction following (low difficulty)**
the instruction is a straightforward mathematical request. the difficulty lies not in understanding *what* to do, but in the *execution* of the mathematical derivation.

\#\#\#\# **5. coding/agentic (n/a)**
unless the model decides to use a python interpreter (like sympy) to solve the sum, this is a purely symbolic reasoning task. if the model uses a tool, the difficulty shifts to correctly formulating the sympy code to handle the summation of binomial products.

---

\#\#\# **critical success factors \& potential failure points**

*   **critical success factor:** the ability to relate the sum to the identity $\sum \binom{2k}{k}\binom{2n-2k}{n-k} = 4^n$ and then systematically apply the $(2k+1)^5$ weight.
*   **potential failure point (algebraic drift):** the expansion of $(2k+1)^5$ creates many terms. a model is likely to make a sign error or a coefficient error during the expansion or the subsequent simplification.
*   **potential failure point (over-simplification):** the model might attempt to use a general formula for $\sum k^p \binom{2k}{k}\binom{2n-2k}{n-k}$ that it "remembers" incorrectly, leading to a plausible-looking but wrong closed form.

\#\#\# **summary table**

| dimension | difficulty | key challenge |
| :--- | :--- | :--- |
| **reasoning** | very high | handling the 5th-degree polynomial weight via operators or identities. |
| **knowledge retrieval** | high | recalling specific binomial convolution identities and generating functions. |
| **comprehension** | moderate | correct interpretation of latex and "closed form" requirement. |
| **instruction following** | low | simple direct request. |
| **overall** | **high** | requires advanced symbolic manipulation and precision. |
\end{tcolorbox}

\subsection{VDAR-Router Retrieved Result}

\begin{tcolorbox}[casebox, title=Rank 1]

\textbf{Retrieved Query}

Let $\prod_{i=0}^{19}\left( 1+x^{3^i}+x^{2\cdot 3^i}+
x^{3\cdot 3^i}\right) = \sum_k a_kx^k$. Find $\sum_k a_k^2$.

\vspace{0.35em}
\textbf{Retrieved Difficulty Analysis}

**Query Difficulty Analysis Summary**

**Overall Difficulty Level: HIGH**

This mathematical problem presents significant challenges for a large language model across multiple dimensions:

**1. Reasoning (HIGH)**
- Requires multi-step mathematical insight to connect polynomial coefficients to combinatorial counting
- Needs recognition of the identity $\sum_k a_k^2 = [x^0]P(x)P(x^{-1})$
- Involves understanding modified base-3 representation structure where digits range from 0-3
- The mathematical derivation is non-obvious and requires competition-level insight

**2. Comprehension (MODERATE-HIGH)**
- Must parse complex mathematical notation with nested products, exponents, and summations
- Requires understanding the conceptual meaning of polynomial coefficients in this context
- The relationship between the given expression and target quantity is not immediately apparent

**3. Knowledge Retrieval (HIGH)**
- Needs specialized knowledge of polynomial algebra and coefficient properties
- Requires access to Parseval-like identities for coefficient sums
- Must understand combinatorial counting techniques in polynomial products
- This is advanced mathematical knowledge beyond general training data

**4. Instruction Following (LOW)**
- Query is clearly stated with well-defined objectives
- No ambiguity in what needs to be computed

**5. Agentic/Coding/Multilingual (LOW)**
- No tool use, planning, or external information gathering required
- Pure mathematical derivation with no code generation
- Single language with standard mathematical notation

**Key Challenge**: The primary difficulty is the mathematical reasoning and knowledge required to recognize and apply the correct mathematical identity. Once the insight is found, the computation is straightforward ($4^{20}$), but finding that insight requires sophisticated mathematical thinking at an advanced competition level.

**Estimated Difficulty**: This represents an AIME/IMO-level competition mathematics problem that would challenge most general-purpose LLMs without specialized mathematical training or reasoning capabilities.
\end{tcolorbox}

\begin{tcolorbox}[casebox, title=Rank 2]
\textbf{Retrieved Query}

Given: $\sigma_2(n)=\sum_{i=1}^ni^2=\sum_{i=1}^n\sum_{j=1}^i(2j-1)$.
This double sum can be interpreted as a single Lebesgue sum: $\sigma_2(n)=\sum_{j=1}^n(2j-1)\mu_2(j).$
determine tha form of the function $\mu_2(j)$

\vspace{0.35em}
\textbf{Retrieved Difficulty Analysis}

**Query Difficulty Analysis Summary**

**Overall Difficulty Level: Moderate to High**

**Key Difficulty Dimensions:**

1. **Mathematical Reasoning (High Difficulty)**: Requires multi-step logical deduction to convert a double summation to a single summation. The model must recognize that for each j from 1 to n, the term $(2j-1)$ appears in the inner sum for all $i \geq j$, giving $\mu_2(j) = (n - j + 1)$. This requires understanding summation order manipulation and counting functions.

2. **Mathematical Comprehension (Moderate Difficulty)**: Requires correct parsing of LaTeX notation and understanding the relationship between the three equation forms. The term "Lebesgue sum" in discrete context may cause confusion as it typically refers to continuous measure theory.

3. **Knowledge Retrieval (Moderate Difficulty)**: Requires specialized knowledge of summation manipulation techniques, discrete mathematics, and combinatorial counting principles not commonly found in general training data.

4. **Instruction Following (Low Difficulty)**: The instruction is clear and direct - determine the form of $\mu_2(j)$ - but requires mathematical derivation rather than simple information extraction.

5. **Pattern Recognition (Moderate Difficulty)**: Requires recognizing the counting pattern in the nested summation structure where each j value appears multiple times across different i values.

**Critical Success Factors:**
- Correct interpretation of double summation limits
- Accurate counting of term occurrences across nested sums
- Understanding of discrete counting functions in summation contexts

**Potential Failure Points:**
- Misinterpreting the relationship between i and j in the double sum
- Confusing continuous Lebesgue measure with discrete counting weights
- Arithmetic errors in deriving the counting function $\mu_2(j)$

**Expected Solution:** $\mu_2(j) = (n - j + 1)$ for $j = 1, 2, ..., n$
\end{tcolorbox}

\begin{tcolorbox}[casebox, title=Rank 3]
\textbf{Retrieved Query}

Find the sum of all integers $k$ such that $\binom{23}{4} + \binom{23}{5} = \binom{24}{k}$.

\vspace{0.35em}
\textbf{Retrieved Difficulty Analysis}

The query presents a **medium difficulty** challenge for a Large Language Model, primarily driven by **Knowledge Retrieval** and **Reasoning** dimensions.

**Key Difficulty Factors:**
1.  **Specialized Knowledge:** Requires specific recall of combinatorial identities, specifically Pascal's Identity and the Symmetry Property of Binomial Coefficients. This is not general knowledge but domain-specific mathematical curriculum.
2.  **Reasoning Complexity:** The task involves multi-step symbolic manipulation. The critical challenge is recognizing that the equation yields **two** valid integer solutions for $k$ due to symmetry ($\binom{n}{k} = \binom{n}{n-k}$), rather than just one. A model must avoid the common bias of assuming a unique solution.
3.  **Constraint Verification:** The model must ensure the identified integers fall within the valid domain ($0 \le k \le n$) and sum them correctly.
4.  **Instruction Following:** The meta-task requires analyzing the difficulty without directly solving the problem, demanding a clear separation between the analysis process and the mathematical solution.

**Overall Assessment:**
The query tests an LLM's ability to apply precise mathematical rules and handle edge cases in symmetry, rather than simple pattern matching. Success depends on accurate symbolic reasoning and comprehensive domain knowledge retrieval.
\end{tcolorbox}

\subsection{KNN-Router Retrieved Result}

\begin{tcolorbox}[casebox, title=Rank 1]
\textbf{Retrieved Query}

Find the sum of all integers $k$ such that $\binom{23}{4} + \binom{23}{5} = \binom{24}{k}$.
\end{tcolorbox}

\begin{tcolorbox}[casebox, title=Rank 2]
\textbf{Retrieved Query}

find the function f with the lowest complexity such that there exists a constant C>0 with 
$\sum_{m=0}^n (-1)^m {n\choose m} \frac{\Gamma(\frac{3}{2}+n)}{\Gamma(\frac{3}{2}+n-m)}\leq Cf(n)$ for all $n\in N$
\end{tcolorbox}

\begin{tcolorbox}[casebox, title=Rank 3]
\textbf{Retrieved Query}

Find a closed expression of the infinite product $\prod_{n=0}^{\infty}(1-e^{-(2n+1)\pi})$.
\end{tcolorbox}

%% file: tables/06-modellist.tex
\begin{table*}[htbp]
\centering
\begin{tabular}{p{3cm}p{12cm}}
\hline
\textbf{Benchmark} & \textbf{Models} \\
\hline
RouterBench & WizardLM/WizardLM-13B-V1.2, claude-instant-v1, claude-v1, claude-v2, gpt-3.5-turbo-1106, gpt-4-1106-preview, meta/code-llama-instruct-34b-chat, meta/llama-2-70b-chat, mistralai/mistral-7b-chat, mistralai/mixtral-8x7b-chat, zero-one-ai/Yi-34B-Chat \\
\hline
LLMRouterBench & claude-sonnet-4, deepseek-r1-0528, deepseek-v3-0324, deepseek-v3.1-terminus, gemini-2.5-flash, gemini-2.5-pro, glm-4.6, gpt-5, intern-s1, kimi-k2-0905, qwen3-235b-a22b-2507 \\
\hline
ArenaExpert5K & MiMo-7B, MiMo-VL-7B-RL-2508, amazon-nova-experimental-chat-05-14, amazon.nova-pro-v1:0, chatgpt-4o-latest-20250326, chatgpt-4o-latest-20250326-old, claude-3-5-haiku-20241022, claude-3-5-sonnet-20241022, claude-3-7-sonnet-20250219, claude-3-7-sonnet-20250219-thinking-32k, claude-opus-4-1-20250805, claude-opus-4-1-20250805-thinking-16k, claude-opus-4-1-20250805-thinking-16k-old, claude-opus-4-20250514, claude-opus-4-20250514-thinking-16k, claude-sonnet-4-20250514, claude-sonnet-4-20250514-thinking-32k, claude-sonnet-4-5-20250929-old, claude-sonnet-4-5-20250929-thinking-32k, command-a-03-2025, deepseek-r1-0528, deepseek-v3-0324, deepseek-v3.1, deepseek-v3.1-terminus, deepseek-v3.1-terminus-thinking, deepseek-v3.1-thinking, deepseek-v3.2-exp, deepseek-v3.2-exp-thinking, gemini-2.0-flash-001, gemini-2.0-flash-thinking-exp-01-21, gemini-2.5-flash-lite-preview-06-17-thinking, gemini-2.5-flash-lite-preview-09-2025-no-thinking, gemini-2.5-flash-preview-04-17, gemini-2.5-flash-preview-09-2025, gemini-2.5-pro-preview-03-25, gemini-2.5-pro-preview-05-06, gemma-3-27b-it, gemma-3n-e4b-it, glm-4.5, glm-4.5-air, glm-4.5v, glm-4.6, gpt-4.1-2025-04-14, gpt-4.1-mini-2025-04-14, gpt-4o-2024-11-20, gpt-4o-mini-2024-07-18, gpt-5-chat, gpt-5-high, gpt-5-high-new-system-prompt, gpt-5-mini-high, gpt-5-nano-high, gpt-5-old, gpt-oss-120b, gpt-oss-20b, grok-3-mini-beta, grok-3-mini-high, grok-3-preview-02-24, grok-4-0709, grok-4-0709-old2, grok-4-fast, grok-4-fast-reasoning, hunyuan-t1-20250711, hunyuan-turbos-20250416, hunyuan-vision-1.5-thinking, ibm-granite-h-small, kimi-k2-0711-preview, kimi-k2-0905-preview, ling-flash-2.0, llama-3.3-70b-instruct, llama-4-maverick-03-26-experimental, llama-4-maverick-17b-128e-instruct, llama-4-scout-17b-16e-instruct, longcat-flash-chat, magistral-medium-2506, mai-1-preview, minimax-m1, mistral-medium-2505, mistral-medium-2508, mistral-small-2506, mistral-small-3.1-24b-instruct-2503, nvidia-llama-3.3-nemotron-super-49b-v1.5, o3-2025-04-16, o3-mini, o4-mini-2025-04-16, qwen-max-2025-01-25, qwen-vl-max-2025-08-13, qwen3-235b-a22b, qwen3-235b-a22b-instruct-2507, qwen3-235b-a22b-instruct-2507-invalid, qwen3-235b-a22b-no-thinking, qwen3-235b-a22b-thinking-2507, qwen3-30b-a3b, qwen3-30b-a3b-instruct-2507, qwen3-coder-480b-a35b-instruct, qwen3-max-2025-09-23, qwen3-max-2025-09-26, qwen3-max-preview, qwen3-next-80b-a3b-instruct, qwen3-next-80b-a3b-thinking, qwen3-vl-235b-a22b-instruct, qwen3-vl-235b-a22b-thinking, qwq-32b, ring-flash-2.0, step-1o-turbo-202506, step-3 \\
\hline
\end{tabular}
\caption{Candidate Models across each datasets.}
\label{tab:model_list}
\end{table*}